\documentclass{article}
\usepackage{flushend} 
\usepackage{balance}
 \usepackage{array,multirow,graphicx}
\usepackage{float}

\usepackage{microtype}
\usepackage{graphicx}
\usepackage{subfigure}
\usepackage{booktabs} 

\usepackage{hyperref}

\usepackage[accepted]{sty}
\usepackage{graphicx}
\usepackage{pythonhighlight}

\usepackage{times}
\usepackage{helvet}
\usepackage{courier}
\usepackage{subfigure} 
\usepackage{cancel}
\RequirePackage{amsthm,amsmath}
\usepackage{graphicx, amssymb, mathrsfs,amsfonts,url, bm}

\newcommand{\real}{\mathbb{R}}

\newcommand{\gap}{\,\,\,\,\,\,\,\,}

\newcommand{\btheta}{\boldsymbol\theta}

\newcommand{\ba}{\bm{a}}
\newcommand{\bA}{\bm{A}}

\newcommand{\bC}{\bm{C}}

\newcommand{\bE}{\bm{E}}

\newcommand{\bI}{\bm{I}}

\newcommand{\br}{\bm{r}}

\newcommand{\bW}{\bm{W}}
\newcommand{\bx}{\bm{x}}
\newcommand{\bX}{\bm{X}}
\newcommand{\by}{\bm{y}}
\newcommand{\bY}{\bm{Y}}

\newcommand{\bZ}{\bm{Z}}

\newcommand{\bzero}{\mathbf{0}}

\newcommand{\gammadist}{\mathcal{G}}
\newcommand{\inversegammadist}{\mathcal{G}^{-1}}

\newcommand{\normal}{\mathcal{N}}
\newcommand{\truncatednormal}{\mathcal{TN}}
\newcommand{\generaltruncatednormal}{\mathcal{GTN}}

\newcommand{\gtnsng}{\mathcal{GTNSNG}}

\usepackage[font=small,skip=2pt]{caption}
 
\newcommand{\titlethis}{Bayesian Low-Rank Interpolative Decomposition for Complex Datasets}

\icmltitlerunning{\titlethis}

\begin{document}

\twocolumn[
\icmltitle{\titlethis}

\begin{icmlauthorlist}
	\icmlauthor{\parbox{10em}{\gap \gap Jun Lu\\ {jun.lu.locky@gmail.com}}}{te}
\end{icmlauthorlist}

\icmlaffiliation{te}{Correspondence to: Jun Lu $<$jun.lu.locky@gmail.com$>$. Copyright 2022 by the author(s)/owner(s). May 30th, 2022}



\vskip 0.3in
]



\printAffiliationsAndNotice{}  

\begin{abstract}
In this paper, we introduce a probabilistic model for learning interpolative decomposition (ID), which is commonly used for feature selection, low-rank approximation, and identifying hidden patterns in data, where the matrix factors are latent variables associated with each data dimension. 
Prior densities with support on the specified subspace are used to address the constraint for the magnitude of the factored component of the observed matrix.
Bayesian inference procedure based on Gibbs sampling is employed. We evaluate the model on a variety of real-world datasets including CCLE $EC50$, CCLE $IC50$, CTRP $EC50$, and MovieLens 100K datasets with different sizes, and dimensions, and show that the proposed Bayesian ID GBT and GBTN models lead to smaller reconstructive errors compared to existing randomized approaches.


\end{abstract}
\section{Introduction}
Matrix factorization methods such as singular value decomposition (SVD), factor analysis, principal component analysis (PCA), and independent component analysis (ICA) have been used extensively over the years to reveal hidden structures of matrices in many fields of science and engineering such as deep learning \citep{liu2015sparse}, recommendation systems \citep{comon2009tensor, lu2021numerical, lu2022matrix}, computer vision \citep{goel2020survey}, clustering and classification \citep{li2009non, wang2013non, lu2021survey}, collaborative filtering \citep{marlin2003modeling, lim2007variational, mnih2007probabilistic, raiko2007principal, chen2009collaborative, brouwer2017prior, lu2022matrix} and machine learning in general \citep{lee1999learning}. 



Moreover, low-rank approximations are essential in modern data science. 
Low-rank matrix approximation with respect to the Frobenius norm - minimizing the sum squared differences to the target matrix - can be easily solved with
singular value decomposition. For many applications, however, it is sometimes advantageous to work with a basis that consists of a subset of the columns of the observed matrix itself \citep{halko2011finding, martinsson2011randomized}. 
The interpolative decomposition (ID) provides one such approximation. Its distinguishing feature is that it reuses columns from the original matrix. This enables it to preserve matrix properties such as sparsity and nonnegativity that also help save space in memory.

Interpolative decomposition is widely used as a feature selection tool that extracts the essence and
allows dealing with big data which is originally too large
to fit into the RAM. In addition, it removes the non-relevant
parts of the data which consist of error and redundant information \citep{liberty2007randomized, halko2011finding, martinsson2011randomized, ari2012probabilistic, lu2021numerical}.
In the meantime, finding the indices associated with the spanning columns is frequently valuable for the purpose of data interpretation and analysis, it can be very useful to identify a subset of the columns that distills the information in the matrix. If the columns of the observed matrix have some specific interpretations, e.g., they are transactions in a transaction dataset, then the columns of the factored matrix in ID will have the same meaning as well.
The factored matrix obtained from interpolative decomposition is also numerically stable since its maximal magnitude is limited to a certain range.

On the other hand, matrix factorizations can also be
thought of as statistical models in which we seek the factorization that offers the maximum marginal likelihood (MML)
for the underlying data \citep{ari2012probabilistic, brouwer2017prior}. Probabilistic interpretations are
investigated for many popular matrix decompositions in the
literature such as  real-valued matrix factorization, nonnegative matrix factorization (NMF)
\citep{brouwer2017prior, lu2022flexible}, principal component analysis \citep{tipping1999probabilistic}, and generalized to tensor factorizations \citep{schmidt2009probabilistic}. In the meantime, probabilistic models can easily accommodate constraints on the specific range of the factored matrix.

\begin{figure*}[h]
	\centering  
	\subfigtopskip=2pt 
	\subfigbottomskip=9pt 
	\subfigcapskip=-5pt 
	\includegraphics[width=0.95\textwidth]{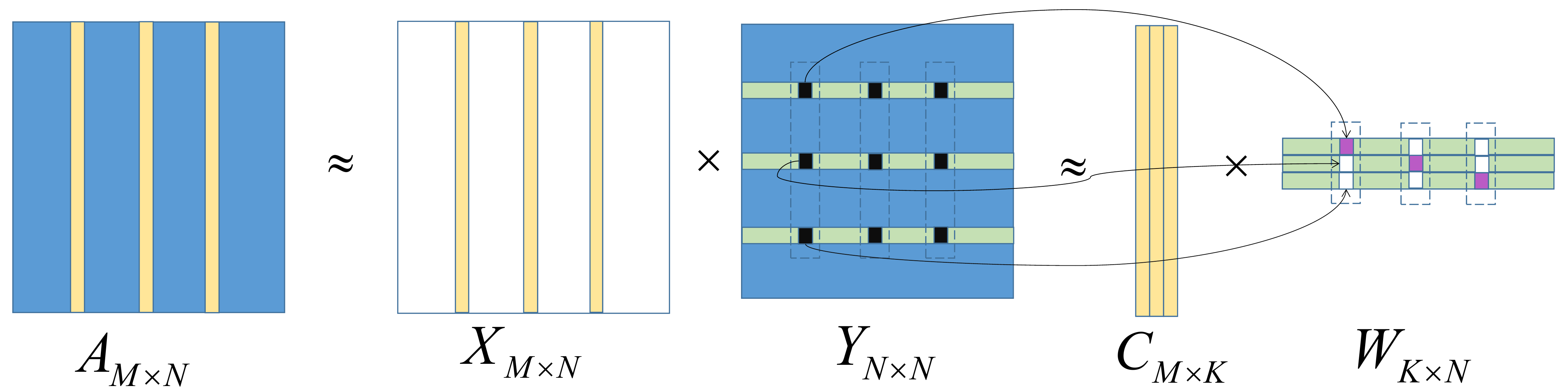}
\caption{Demonstration of the interpolative decomposition of a matrix where the yellow vector denotes
the basis columns of matrix $\bA$, white entries denote zero, purple entries denote
one, blue and black entries denote elements that are not necessarily zero. The Bayesian ID models find the approximation $\bA\approx\bX\bY$ and the post processing procedure finds the approximation $\bA\approx\bC\bW$.}
\label{fig:id-column}
\end{figure*}

In this light, we focus on the Bayesian ID (BID) of underlying matrices.
The ID problem of observed matrix $\bA$ can be stated as $\bA=\bC\bW+\bE$, where $\bA= [\ba_1, \ba_2, \ldots, \ba_N]\in \real^{M\times N}$ is approximately factorized into an $M\times K$ matrix $\bC\in \real^{M\times K}$ containing $K$ basis columns of $\bA$ and a $K\times N$ matrix $\bW\in \real^{K\times N}$ with entries no larger than 1 in magnitude \footnote{A weaker construction is to assume no entry of $\bY$ has an absolute value greater than 2. See proof of the existence of the decomposition in \citet{lu2021numerical}.}; the noise is captured by matrix $\bE\in \real^{M\times N}$. 
Training such models amounts to finding the best rank-$K$ approximation to the observed $M\times N$ target matrix $\bA$ under the given loss function. Let $\br\in \{0,1\}^N$ be the \textit{state vector} with each element indicating the type of the corresponding column, basis column or interpolated (remaining) column: if $r_n=1$, then the $n$-th column $\ba_n$ is a basis column; if $r_n=0$, then $\ba_n$ is interpolated using the basis columns plus some error term. Suppose further $J$ is the set of the indices of the selected basis columns, $I$ is the set of the indices of the interpolated columns such that $J \cup I =\{1,2,\ldots, N\}$, $J=J(\br)=\{n|r_n=1\}_{n=1}^N$, and $I=I(\br)=\{n|r_n=0\}_{n=1}^N$. Then $\bC$ can be described as $\bC=\bA[:,J]$ where the colon operator implies all indices. The approximation $\bA\approx \bC\bW$ can be equivalently stated that $\bA\approx\bC\bW=\bX\bY$ where $\bX\in \real^{M\times N}$ and $\bY\in \real^{N\times N}$ such that $\bX[:,J]=\bC$, $\bX[:,I] = \bzero\in \real^{M\times (N-K)}$; $\bW = \bY[J,:]$. 
We also notice that there exists an identity matrix $\bI\in \real^{K\times K}$ in $\bW$ and $\bY$:
\begin{equation}\label{equation:submatrix_bid_identity}
\bI = \bW[:,J] = \bY[J,J].
\end{equation}
To find the low-rank ID of $\bA\approx\bC\bW$ then can be transformed into the problem of finding the $\bA\approx\bX\bY$ with state vector $\br$ recovering the submatrix $\bC$ (see Figure~\ref{fig:id-column}).
To evaluate the approximation, \textit{reconstruction error} measured by mean squared error (MSE or Frobenius norm) is minimized (assume $K$ is known):
\begin{equation}\label{equation:idbid-per-example-loss}
\mathop{\min}_{\bW,\bZ} \,\, \frac{1}{MN}\sum_{n=1}^N \sum_{m=1}^{M} \left(a_{mn} - \bx_m^\top\by_n\right)^2,
\end{equation}
where $\bx_m$, $\by_n$ are the $m$-th row and $n$-th column of $\bX$, $\bY$ respectively.
In this paper, we approach the magnitude constraint in $\bW$ and $\bY$ by considering the Bayesian ID model as a latent factor model and we describe a fully specified graphical model for the problem and employ Bayesian learning methods to infer the latent factors. In this sense, explicit magnitude constraints are not required on the latent factors, since this is naturally taken care of by the appropriate choice of prior distribution; here we use general-truncated-normal prior.

The main contribution of this paper is to propose a novel Bayesian ID method which is flexible called the \textit{GBT} model. 
To further favor flexibility and insensitivity on the hyperparameter choices, we also propose the hierarchical model known as the \textit{GBTN} algorithm, which has simple conditional density forms with little extra computation, 
Meanwhile, the method is easy to implement. We show that our method can be successfully applied to the large, sparse, and very imbalanced Movie-User dataset, containing 100,000 user/movie ratings.

\section{Related Work}
The most popular algorithm to compute the low-rank ID approximation is the randomized ID (RID) algorithm \citep{liberty2007randomized}.  At a high level, the algorithm randomly samples $M>K$ columns from $\bA$, uses column-pivoted QR \citep{lu2021numerical} to select $K$ of those $M$ columns for basis matrix $\bC$, and then computes $\bW$ via least squares. $M$ is usually set to $M=1.2K$ to oversample the columns so as to capture a large portion of the range of matrix $\bA$.

We propose a new Bayesian approach for interpolative decomposition, which is designed to have a strict constraint on the magnitude of the factored matrix $\bW$. 
The proposed GBT and GBTN models introduce a general-truncated-normal density over the factored matrix $\bW$
that finds interpolated columns to reconstruct the observed matrix.

\subsection{Probability Distributions}\label{section:probability_distribution}
We introduce all notations and probability distribution in this section.

$\normal(x|\mu, \tau^{-1}) =\sqrt{\frac{ \tau}{2\pi}}\exp\{-\frac{\tau}{2} (x-\mu)^2\}$ is a Gaussian distribution with mean $\mu$ and precision $\tau$ (variance $\sigma^2=\tau^{-1}$).

$\gammadist(x|\alpha, \beta)= \frac{\beta^\alpha}{\Gamma(\alpha)} x^{\alpha-1}\exp\{-\beta x\}u(x)$ is a Gamma distribution where $\Gamma(\cdot)$ is the gamma function and $u(x)$ is the unit step function that has a value of $1$ when $x\geq0$ and 0 otherwise.

$\inversegammadist(x|\alpha, \beta)= \frac{\beta^\alpha}{\Gamma(\alpha)} x^{-\alpha-1}\exp\{-\frac{\beta}{x}\}u(x)$ is an inverse-Gamma distribution.


$\truncatednormal(x|\mu,\tau^{-1}) =\frac{\sqrt{\frac{\tau}{2\pi}} \exp\{-\frac{\tau}{2} (x-\mu)^2 \} } 
{1-\Phi(-\mu\sqrt{\tau})} u(x)$
is a truncated-normal (TN) with zero density below $x=0$ and renormalized to integrate to one. $\mu$ and $\tau$ are known as the ``parent mean" and ``parent precision". $\Phi(\cdot)$ is the cumulative distribution function of standard normal density $\normal(0,1)$.


$\generaltruncatednormal(x|\mu, \frac{1}{\tau}, a, b)=\frac{\sqrt{\frac{\tau}{2\pi}} \exp \{-\frac{\tau}{2}(x-\mu)^2  \}  }{\Phi((b-\mu)\cdot \sqrt{\tau})-\Phi((a-\mu)\cdot \sqrt{\tau})}$$u(x|a,b)$
is a general-truncated-normal (GTN) with zero density below $x=a$ or above $x=b$ and renormalized to integrate to one where $u(x|a,b)$ is a step function that has a value of 1 when $a\leq x\leq b$ and 0 otherwise. Similarly, $\mu$ and $\tau$ are known as the ``parent mean" and ``parent precision" of the normal distribution. When $a=0$ and $b=\infty$, the GTN distribution reduces to a TN density.

\section{Bayesian Interpolative Decomposition}

\subsection{Bayesian GBT and GBTN Models for ID}

\begin{figure}[h]
\centering  
\vspace{-0.35cm} 
\subfigtopskip=2pt 
\subfigbottomskip=6pt 
\subfigcapskip=-2pt 
\subfigure[GBT.]{\includegraphics[width=0.231\textwidth]{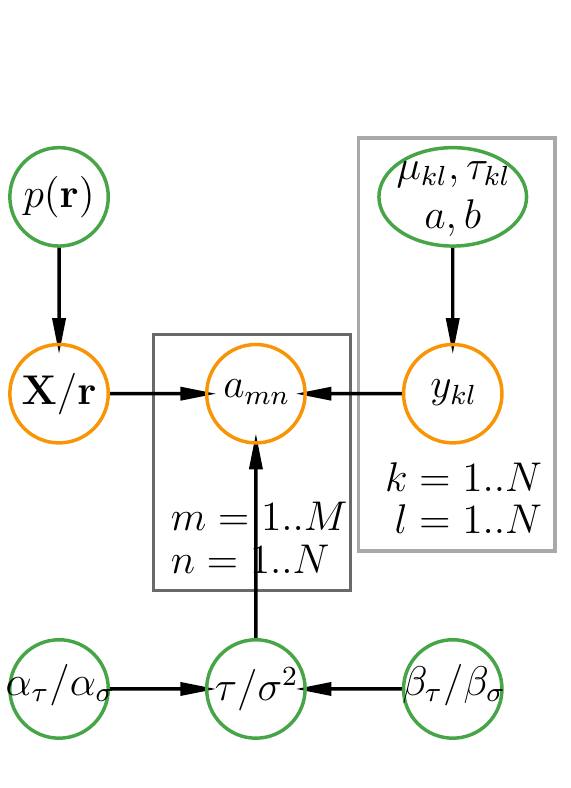} \label{fig:bmf_bid_GBT}}
\subfigure[GBTN.]{\includegraphics[width=0.231\textwidth]{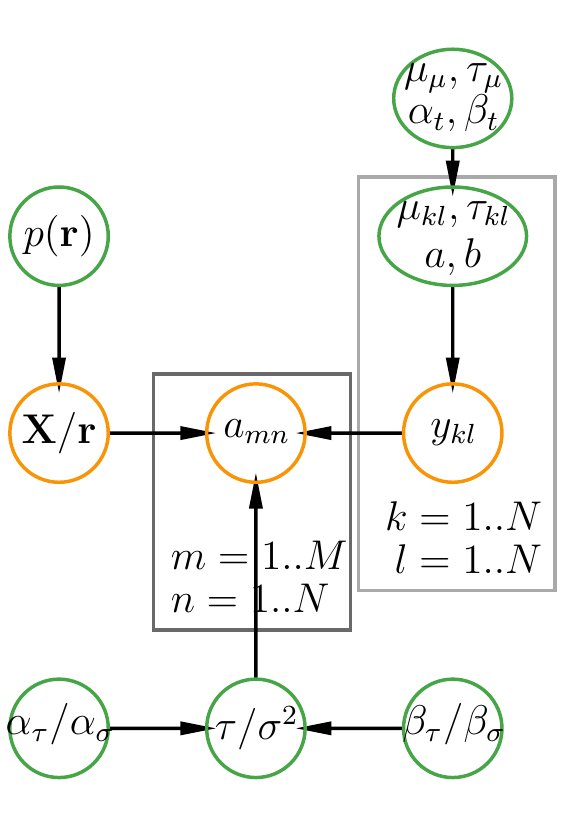} \label{fig:bmf_bid_GBTN}}
\caption{Graphical representation of GBT and GBTN models. Orange circles represent observed and latent variables, green circles denote prior variables, and plates represent repeated variables. ``/" in the variable represents ``or", and comma ``," in the variable represents ``and". Parameters $a,b$ are fixed with $a=-1,b=1$ in our case; while a weaker construction can set them to $a=-2,b=2$.}
\label{fig:bmf_bids}
\end{figure}

We view the data $\bA$ as being produced according to the probabilistic generative process shown in Figure~\ref{fig:bmf_bids}. The observed $(m,n)$-th entry $a_{mn}$ of matrix $\bA$ is modeled using a Gaussian likelihood function with variance $\sigma^2$ and mean given by the latent decomposition $\bx_m^\top\by_n$ (Eq.~\eqref{equation:idbid-per-example-loss}),
\begin{equation}\label{equation:grrn_data_entry_likelihood}
p(a_{mn} | \bx_m^\top\by_n, \sigma^2) = \normal(a_{mn}|\bx_m^\top\by_n, \sigma^2).
\end{equation}
We choose a conjugate prior over the data variance, an inverse-Gamma distribution with shape $\alpha_\sigma$ and scale $\beta_\sigma$, 
\begin{equation}\label{equation:prior_grrn_gamma_on_variance}
	p(\sigma^2 | \alpha_\sigma, \beta_\sigma) = \inversegammadist(\sigma^2 | \alpha_\sigma, \beta_\sigma).
\end{equation}
While it can also be equivalently given a conjugate Gamma prior over the precision and we shall not repeat the details.

We treat the latent variables $y_{kl}$'s as random variables. And we need prior densities over these latent variables to express beliefs for their values, e.g., constraint with magnitude smaller than 1 in this context though there are many other constraints (nonnegativity in \citet{lu2022flexible}, semi-nonnegativity in \citet{ding2008convex}, discreteness in \citet{gopalan2014bayesian, gopalan2015scalable}).
Here we assume further that the latent variable $y_{kl}$'s are independently drawn from a general-truncated-normal prior
\begin{equation}\label{equation:rn_prior_bidd}
p(y_{kl} | \cdot ) = \generaltruncatednormal(y_{kl} | \mu_{kl}, (\tau_{kl})^{-1}, a=-1, b=1).
\end{equation}
This prior serves to enforce the constraint on the components $\bY$ with no entry of $\bY$ having an absolute value greater than 1, and is conjugate to the Gaussian likelihood. The posterior density is also a general-truncated-normal distribution. While in a weaker construction of interpolative decomposition, the constraint on the magnitude can be loosened to 2; the prior is flexible in that the parameters can be then set to $a=-2, b=2$ accordingly. 

\paragraph{Hierarchical prior}
To further favor flexibility, we choose a convenient joint hyperprior density over the parameters $\{\mu_{kl}, \tau_{kl}\}$ of GTN prior in Eq.~\eqref{equation:rn_prior_bidd}, namely, the GTN-scaled-normal-Gamma (GTNSNG) prior,
\begin{equation}
\begin{aligned}
&\gap p(\mu_{kl}, \tau_{kl} |\cdot) 
=\gtnsng(\mu_{kl}, \tau_{kl}| \mu_\mu, \frac{1}{\tau_\mu},\alpha_t, \beta_t)\\
&=	\big\{\Phi((b-\mu_\mu)\cdot \sqrt{\tau_\mu})-\Phi((a-\mu_\mu)\cdot \sqrt{\tau_\mu})\big\}\\
&\gap \cdot 
\normal(\mu_{kl}| \mu_\mu, (\tau_\mu)^{-1}) \cdot \gammadist(\tau_{kl} | a_t, b_t).
\end{aligned}
\end{equation}
This prior can decouple parameters $\mu_{kl}, \tau_{kl}$, and the posterior conditional densities of them are normal, and Gamma respectively due to this convenient scale.

\paragraph{Terminology} There are three types of choices we make that determine
the specific type of matrix decomposition model we use, namely, the likelihood function, the priors we place over the factor matrices $\bW$ and $\bZ$, and whether we use any further hierarchical priors. 
We will call the model by the density function in the order of the types of likelihood and priors. For example, if the likelihood function for the model is chosen to be a Gaussian density, and the two prior density functions are selected to be exponential density and Gaussian density functions respectively, then the model will be denoted as Gaussian Exponential-Gaussian (GEG) model. Sometimes, we will put a hyperprior over the parameters of the prior density functions, e.g., we put a Gamma prior over the Gaussian density, then it will further be termed as a Gaussian Exponential-Gaussian Gamma (GEGA) model. In this sense, the proposed hierarchical model is named the \textit{GBT} and \textit{GBTN} model where $B$ stands for Beta-Bernoulli density intrinsically.

\subsection{Gibbs Sampler}

\begin{algorithm}[tb] 
\caption{Gibbs sampler for GBT and GBTN ID models. The procedure presented here may not be efficient but is explanatory. A more efficient one can be implemented in a vectorized manner. By default, uninformative priors are $a=-1, b=1,\alpha_\sigma=0.1, \beta_\sigma=1$, ($\{\mu_{kl}\}=0, \{\tau_{kl}\}=1$) for GBT, ($\mu_\mu =0$, $\tau_\mu=0.1, \alpha_t=\beta_t=1$) for GBTN.} 
	\label{alg:gbtn_gibbs_sampler}  
	\begin{algorithmic}[1] 
\FOR{$t=1$ to $T$}
\STATE Sample state vector $\br$ from Eq.~\eqref{equation:postrerior_gbt_rvector222};
\STATE Update matrix $\bX$ by $\bA[:,J]$ where index vector $J$ is the index of $\br$ with value 1 and set $\bX[:,I]=\bzero$ where index vector $I$ is the index of $\br$ with value 0;
\STATE Sample $\sigma^2$ from $p(\sigma^2 | \bX,\bY, \bA)$ in Eq.~\eqref{equation:posterior_gnt_sigma2}; 
\FOR{$k=1$ to $N$} 
\FOR{$l=1$ to $N$} 
\STATE Sample $y_{kl}$ from Eq.~\eqref{equation:posterior_gbt_ykl};
\STATE (GBTN only) Sample $\mu_{kl}$ from Eq.~\eqref{equation:posterior_gbt_mukl};
\STATE (GBTN only) Sample $\tau_{kl}$ from Eq.~\eqref{equation:posterior_gbt_taukl};
\ENDFOR
\ENDFOR
\STATE Report loss in Eq.~\eqref{equation:idbid-per-example-loss}, stop if it converges.
\ENDFOR
\STATE Report mean loss in Eq.~\eqref{equation:idbid-per-example-loss} after burn-in iterations.
\end{algorithmic} 
\end{algorithm}

In this paper, we use Gibbs sampling since it tends to be very accurate at finding the true posterior. Other than this method, variational Bayesian inference can be an alternative way but we shall not go into the details. We shortly describe the posterior conditional density in this section. A detailed derivation can be found in Appendix~\ref{appendix:gbt_gbtn_derivation}.
The conditional density of $y_{kl}$ is a general-truncated-normal density. Denote all elements of $\bY$ except $y_{kl}$ as $\bY_{-kl}$, we then have
\begin{equation}\label{equation:posterior_gbt_ykl}
\begin{aligned}
&\gap p(y_{kl} | \bA, \bX, \bY_{-kl}, \mu_{kl}, \tau_{kl}, \sigma^2) \\
&\propto  p(\bA|\bX,\bY, \sigma^2) \cdot p(y_{kl}|\mu_{kl}, \tau_{kl} )\\
&=\prod_{i,j=1}^{M,N} \normal \left(a_{ij}| \bx_i^\top\by_j, \sigma^2 \right)
\cdot \generaltruncatednormal(y_{kl} | \mu_{kl}, \frac{1}{\tau_{kl}},-1,1) \\
&
\propto \generaltruncatednormal(y_{kl}| \widetilde{\mu},( \widetilde{\tau})^{-1}, a=-1,b=1),
\end{aligned}
\end{equation}
where $\widetilde{\tau} =\frac{\sum_{i}^{M}  x_{ik} ^2}{\sigma^2} +\tau_{kl}$ is the posterior ``parent precision" of the GTN distribution, and 
$$
\widetilde{\mu} = \bigg(\frac{1}{\sigma^2}  \sum_{i}^{M} x_{ik}  \big(a_{il}-\sum_{j\neq k}^{N}x_{ij}
y_{jl}\big)
+\textcolor{black}{\tau_{kl}\mu_{kl}}
\bigg) \big/ \widetilde{\tau}
$$
is the posterior ``parent mean" of the GTN distribution.

Given the state vector $\br=[r_1,r_2, \ldots, r_N]^\top\in \real^N$, the relation between $\br$
and the index set $J$ is simple; $J = J(\br) = \{n|r_n = 1\}_{n=1}^N$ and $I = I(\br) = \{n|r_n = 0\}_{n=1}^N$. A new value of state vector $\br$ is to select one index $j$ from index set $J$ and another index $i$ from index sets $I$ (we note that $r_j=1$ and $r_i=0$ for the old state vector $\br$) such that 
\begin{equation}\label{equation:postrerior_gbt_rvector}
\begin{aligned}
o_j &= 
\frac{p(r_j=0, r_i=1|\bA,\sigma^2, \bY, \br_{-ji})}
{p(r_j=1, r_i=0|\bA,\sigma^2, \bY, \br_{-ji})}\\
&=
\frac{p(r_j=0, r_i=1)}{p(r_j=1, r_i=0)}\\
&\gap\gap \times
\frac{p(\bA|\sigma^2, \bY, \br_{-ji}, r_j=0, r_i=1)}{p(\bA|\sigma^2, \bY, \br_{-ji}, r_j=1, r_i=0)},
\end{aligned}
\end{equation}
where $\br_{-ji}$ denotes all elements of $\br$ except $j$-th and $i$-th entries.
Trivially, we can set $p(r_j=0, r_i=1)=p(r_j=1, r_i=0)$. Then the full conditionally probability of $p(r_j=0, r_i=1|\bA,\sigma^2, \bY, \br_{-ji})$ can be calculated by 
\begin{equation}\label{equation:postrerior_gbt_rvector222}
p(r_j=0, r_i=1|\bA,\sigma^2, \bY, \br_{-ji}) = \frac{o_j}{1+o_j}.
\end{equation}



Finally, the conditional density of $\sigma^2$ is an inverse-Gamma distribution by conjugacy,
\begin{equation}\label{equation:posterior_gnt_sigma2}
\begin{aligned}
&\gap p(\sigma^2 | \bX, \bY, \bA)
= \inversegammadist(\sigma^2 | \widetilde{\alpha_\sigma}, \widetilde{\beta_\sigma}),
\end{aligned}
\end{equation}
where $\widetilde{\alpha_\sigma} = \frac{MN}{2}+\alpha_\sigma$, 
$\widetilde{\beta_\sigma}=\frac{1}{2} \sum_{i,j=1}^{M,N}(a_{ij}-\bx_i^\top\by_j)^2+\beta_\sigma$.

\paragraph{Extra updates for the GBTN model}
Following the graphical representation of the GBTN model in Figure~\ref{fig:bmf_bids}, the conditional density of $\mu_{kl}$ can be derived,
\begin{equation}\label{equation:posterior_gbt_mukl}
\begin{aligned}
&\gap p(\mu_{kl} | \tau_{kl}, \mu_\mu, \tau_\mu, a_t, b_t, y_{kl})\\
&\propto \generaltruncatednormal(y_{kl} | \mu_{kl}, (\tau_{kl})^{-1}, a=-1, b=1)\\
&\gap \times  \gtnsng(\mu_{kl}, \tau_{kl}| \mu_\mu, (\tau_\mu)^{-1},\alpha_t, \beta_t)\\
&\propto \normal(\mu_{kl}| \widetilde{m},(\,\widetilde{t}\,)^{-1}),
\end{aligned}
\end{equation}
where $\widetilde{t} = \tau_{kl}+\tau_\mu$, and $\widetilde{m} =(\tau_{kl}y_{kl}+\tau_\mu\mu_\mu)/\widetilde{t}$ are the posterior precision and mean of the normal density. Similarly, the conditional density of $\tau_{kl}$ is,
\begin{equation}\label{equation:posterior_gbt_taukl}
\begin{aligned}
	&\gap p(\tau_{kl} | \mu_{kl}, \mu_\mu, \tau_\mu, a_t, b_t, y_{kl})\\
&\propto \generaltruncatednormal(y_{kl} | \mu_{kl}, (\tau_{kl})^{-1}, a=-1, b=1)\\
&\gap	\times  \gtnsng(\mu_{kl}, \tau_{kl}| \mu_\mu, (\tau_\mu)^{-1},\alpha_t, \beta_t)\\
	&\propto \gammadist(\tau_{kl} | \widetilde{a}, \widetilde{b}),
\end{aligned}
\end{equation}
where $\widetilde{a} = a_t+1/2$ and $\widetilde{b}=b_t +  \frac{(y_{kl}- \mu_{kl})^2}{2}$ are the posterior parameters of the Gamma density.
The full procedure is then formulated in Algorithm~\ref{alg:gbtn_gibbs_sampler}.

\subsection{Aggressive Update}
In Algorithm~\ref{alg:gbtn_gibbs_sampler}, we notice that we set $\bX[:,I]=\bzero$ when the new state vector $\br$ is sampled. However, in the next iteration step, the index set $I$ may be updated in which case one entry $i$ of $I$ may be altered to have value of 1: 
$$
r_i=0 \rightarrow r_i=1.
$$
This may cause problems in the update of $y_{kl}$ in Eq.~\eqref{equation:posterior_gbt_ykl}, a zero $i$-th column in $\bX$ cannot update $\bY$ accordingly. One solution is to record a proposal state vector $\br_2$ and a proposal factor matrix $\bX_2$ from the $\br_2$ vector. 
When the update in the next iteration selects the old state vector $\br$, the factor matrix $\bX$ is adopted to finish the updates; while the algorithm chooses the proposal state vector $\br_2$, the proposal factor matrix $\bX_2$ is applied to do the updates. We call this procedure the \textit{aggressive} update.
The \textit{aggressive} sampler for the GBT model is formulated in Algorithm~\ref{alg:gbtn_gibbs_sampler_aggressive}. For the sake of simplicity, we don't include the sampler for the GBTN model because it may be done in a similar way.

\begin{algorithm}[h] 
\caption{\textit{Aggressive} Gibbs sampler for GBT ID model. The procedure presented here may not be efficient but is explanatory. A more efficient one can be implemented in a vectorized manner. By default, uninformative priors are $a=-1, b=1,\alpha_\sigma=0.1, \beta_\sigma=1$, ($\{\mu_{kl}\}=0, \{\tau_{kl}\}=1$) for GBT. 
} 
\label{alg:gbtn_gibbs_sampler_aggressive}  
\begin{algorithmic}[1] 
\FOR{$t=1$ to $T$}
\STATE Sample state vector $\br$ from $\{\br_1, \br_2\}$ by Eq.~\eqref{equation:postrerior_gbt_rvector222};
\STATE Decide $\bY$: $\bY=\bY_1$ if $\br$ is $\br_1$; $\bY=\bY_2$ if $\br$ is $\br_2$;
\STATE Update state vector $\br_1=\br$; 
\STATE Sample proposal state vector $\br_2$ based on $\br$;
\STATE Update matrix $\bX$ by $\br=\br_1$; 
\STATE Update proposal $\bX_2$ by $\br_2$;
\STATE Sample $\sigma^2$ from $p(\sigma^2 | \bX,\bY, \bA)$ in Eq.~\eqref{equation:posterior_gnt_sigma2}; 
\STATE Sample $\bY_1 = \{y_{kl}\}$ using $\bX$;
\STATE Sample $\bY_2 = \{y_{kl}\}$ using $\bX_2$;
\STATE Report loss in Eq.~\eqref{equation:idbid-per-example-loss}, stop if it converges.
\ENDFOR
\STATE Report mean loss in Eq.~\eqref{equation:idbid-per-example-loss} after burn-in iterations.
\end{algorithmic} 
\end{algorithm}

\subsection{Post Processing}
The Gibbs sampling algorithm finds the approximation $\bA\approx \bX\bY$ where $\bX\in \real^{M\times N}$ and $\bY\in \real^{N\times N}$. As stated above, the redundant columns in $\bX$ and redundant rows in $\bY$ can be removed by the index vector $J$: 
$$
\begin{aligned}
\bC&=\bX[:,J]=\bA[:,J],\\
\bW&= \bY[J,:].
\end{aligned}
$$
Since the submatrix $\bY[J,J]=\bW[:,J]$ (Eq.~\eqref{equation:submatrix_bid_identity}) from the Gibbs sampling procedure is not enforced to be an identity matrix (as required in the interpolative decomposition). We need to set it to be an identity matrix manually. This will basically reduce the reconstructive error further. 
The post processing procedure is shown in Figure~\ref{fig:id-column}.


\begin{table}[h]
\begin{tabular}{llll}
\hline
Dataset        & Rows & Columns & Fraction obs. \\ \hline
CCLE $EC50$ & 502 & 48  &0.632\\
CCLE $IC50$ & 504 &48 &0.965\\
CTRP $EC50$ &887 &1090 &0.801\\
MovieLens 100K & 943  & 2946    & 0.072         \\ 
\hline
\end{tabular}
\caption{Overview of the CCLE $EC50$, CCLE $IC50$, CTRP $EC50$, and MovieLens 100K datasets, giving the number of rows, columns (after copying twice for redundancy), and the fraction of entries that are observed. The MovieLens 100K dataset has a larger fraction of unobserved entries.}
\label{table:datadescription}
\end{table}
\begin{figure}[h]
	\centering  
	\subfigtopskip=2pt 
	\subfigbottomskip=2pt 
	\subfigcapskip=-5pt 
\subfigure{\includegraphics[width=0.231\textwidth]{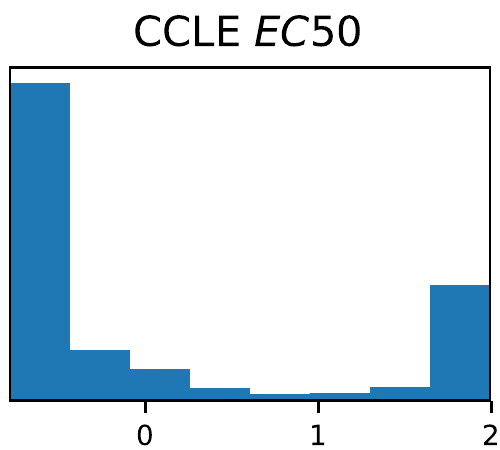} \label{fig:plot_ccle_ec}}
\subfigure{\includegraphics[width=0.231\textwidth]{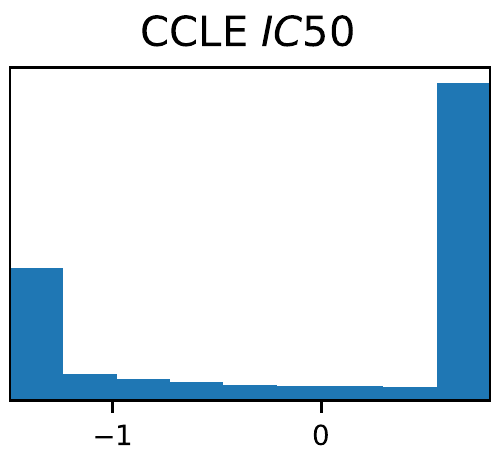} \label{fig:plot_ccle_ic}}
\subfigure{\includegraphics[width=0.231\textwidth]{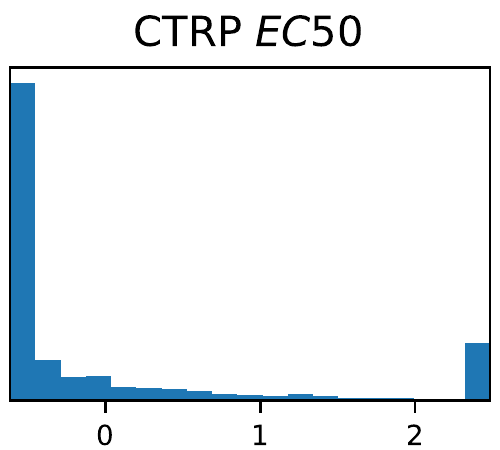} \label{fig:plot_ctrp}}
\subfigure{\includegraphics[width=0.231\textwidth]{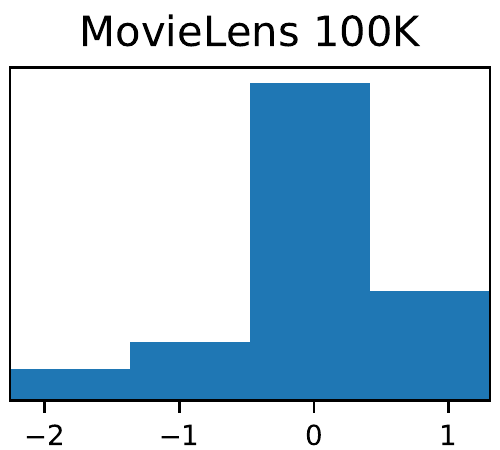} \label{fig:plot_movielens100}}
\caption{Data distribution of CCLE $EC50$, CCLE $IC50$, CTRP $EC50$, and MovieLens 100K datasets. 
The former three datasets have complex distributions and the last one has a Gaussian-like distribution.}
\label{fig:datasets_bids}
\end{figure}
\section{Experiments}\label{section:ader_experiments}

To evaluate the strategy and demonstrate the main advantages of the proposed Bayesian ID method, 
we conduct experiments with different analysis tasks; and different data sets including 
Cancer Cell Line Encyclopedia (CCLE $EC50$ and CCLE $IC50$ datasets, \citep{barretina2012cancer}) and Cancer Therapeutics Response Portal (CRTP $EC50$, \citep{seashore2015harnessing}) from bioinformatics, and MovieLens 100K from movie ratings for different users \citep{harper2015movielens}. Following \citet{brouwer2017prior}, we preprocess these datasets by capping high values to 100 and undoing the natural log transform for the former three datasets. Then we standardize to have zero mean and unit variance for all datasets and filling missing entries by 0. Moreover, the user vectors or movie vectors with less than 3 observed entries are cleaned in the MovieLens 100K dataset. Finally, we copy every column twice in order to increase redundancy in the matrix.
A summary of the four datasets can be seen in Table~\ref{table:datadescription} and their distributions are shown in Figure~\ref{fig:datasets_bids}.
The MovieLens 100K dataset is unbalanced in that it has a larger fraction of unobserved entries with only $7.2\%$ being observed so that the algorithm may tend to simply predict all entries to be zero.


In all scenarios, the same parameter initialization is adopted when conducting different tasks. 
Experimental evidence reveals that post processing procedure can increase performance to a minor level, and that the outcomes of the GBT and GBTN models (aggressive and non-aggressive versions) are relatively similar.
For clarification, we only provide the findings of the GBT model after an aggressive update and post processing.
We compare the results with randomized ID (RID) algorithm \citep{liberty2007randomized}. In a wide range of scenarios across various datasets, the GBT model improves reconstructive error and leads to performances that are as good or better than the randomized ID method in low-rank ID approximation.
We also notice that all the four datasets contain missing entries that are filled by zero values. When the fraction of unobserved data is large, the model can simply predict the entries with zero. So we also report the mean squared error for observed entries, termed as \textit{GBT (Observed)} for the proposed GBT model, and \textit{RID (Observed)} for the randomized ID model.

In order to measure
overall decomposition performance, we use mean square error (MSE, Eq.~\eqref{equation:idbid-per-example-loss}), which measures the similarity between the true and reconstructive; the smaller the better performance.

\subsection{Hyperparameters}
In this experiments, we use  $a=-1, b=1,\alpha_\sigma=0.1, \beta_\sigma=1$, ($\{\mu_{kl}\}=0, \{\tau_{kl}\}=1$) for GBT, ($\mu_\mu =0$, $\tau_\mu=0.1, \alpha_t=\beta_t=1$) for GBTN.
These are very weak prior choices and the models are not sensitive to them.
As long as the hyperparameters are set, the observed or unobserved variables are initialized from random draws as this initialization procedure provides a better initial guess of the right patterns in the matrices.
In all experiments, we run the Gibbs sampler 500 iterations with a burn-in of 100 iterations and a thinning of 5 iterations as the convergence analysis shows the algorithm can converge in less than 50 iterations.

\subsection{Convergence Analysis}
\begin{figure*}[h]
	\centering  
	\subfigtopskip=2pt 
	\subfigbottomskip=2pt 
	\subfigcapskip=-2pt 
	\subfigure[Convergence of the models on the 
	CCLE $EC50$, CCLE $IC50$, CTRP $EC50$, and MovieLens 100K datasets, measuring the training data fit (mean squared error). The algorithm almost converges in less than 50 iterations.]{\includegraphics[width=1\textwidth]{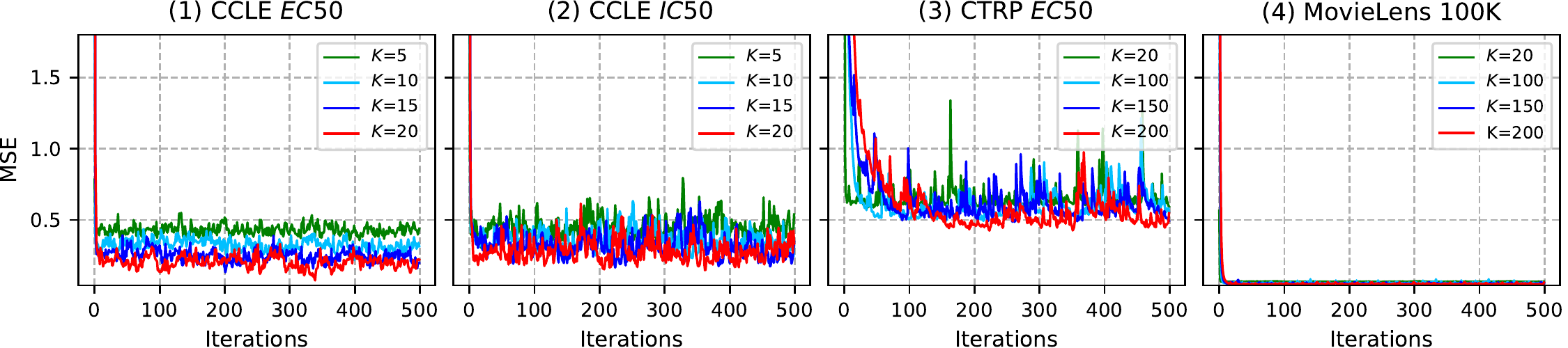} \label{fig:convergences_BIDs}}
	\subfigure[Averated autocorrelation coefficients of samples of $y_{kl}$ computed using Gibbs sampling on the 
	CCLE $EC50$, CCLE $IC50$, CTRP $EC50$, and MovieLens 100K datasets.]{\includegraphics[width=1\textwidth]{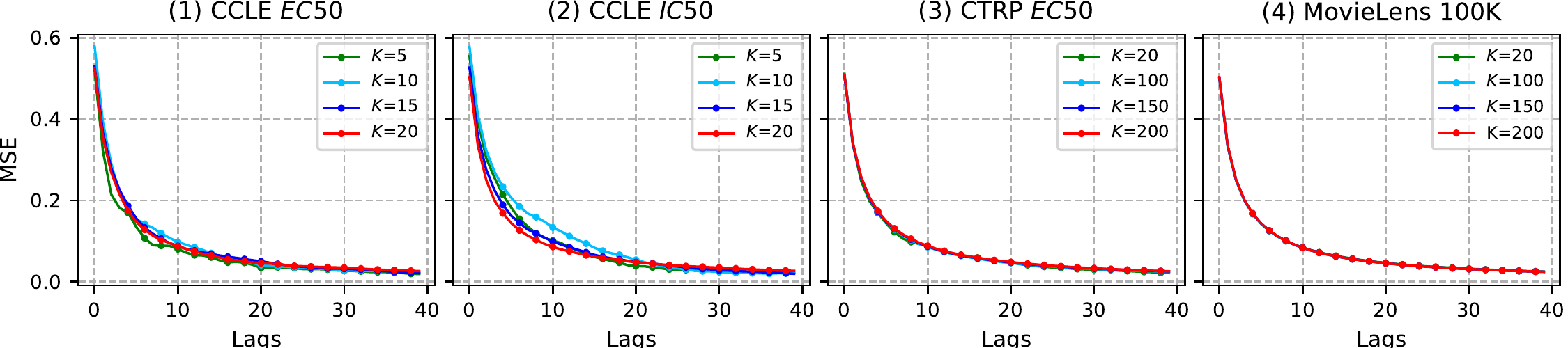} \label{fig:convergences_BIDs_autocorr}}
	\subfigure[Reconstructive analysis on the CCLE $EC50$, CCLE $IC50$, CTRP $EC50$, and MovieLens 100K datasets with increasing latent dimension $K$.]{\includegraphics[width=1\textwidth]{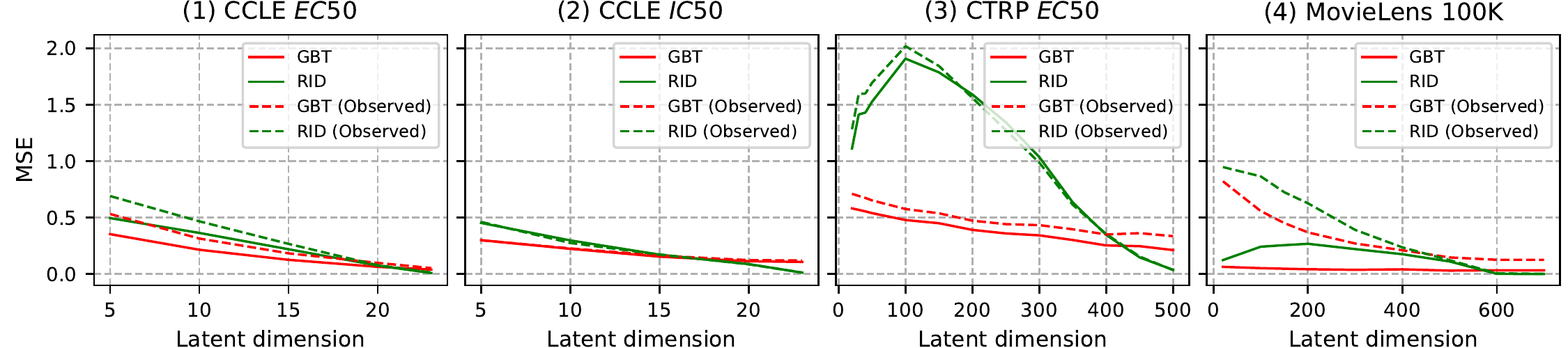} \label{fig:error_bid_all}}
\caption{Convergence results (upper), sampling mixing analysis (middle), and reconstructive results (lower) on the CCLE $EC50$, CCLE $IC50$, CTRP $EC50$, and MovieLens 100K datasets for various latent dimensions.
	}
	\label{fig:allresults_bids}
\end{figure*}

Firstly we show the convergence in terms of iterations on the CCLE $EC50$, CCLE $IC50$, CTRP $EC50$ and MovieLens 100K datasets. We run each model with $K=5, 10, 15, 20$ for the CCLE $EC50$ and CCLE $IC50$ datasets, 
and $K=20, 100, 150, 200$ for the CTRP $EC50$ and MovieLens 100K datasets; and the loss is measured by mean squared error (MSE).
Figure~\ref{fig:convergences_BIDs} shows the average convergence results. Figure~\ref{fig:convergences_BIDs_autocorr} shows autocorrelation coefficients of samples computed using Gibbs sampling. The coefficients are less than 0.1 when the lags are more than 10 showing the mixing of the Gibbs sampler is good. 
In all experiments, the algorithm converges in less than 50 iterations.
On the CCLE $EC50$ and MovieLens 100K datasets, the sampling are less noisy; while on the CCLE $IC50$ and CTRP $EC50$ datasets, the sampling seems to be noisy.  


\begin{table}[h]
	\centering
	\begin{tabular}{c|lrr}
		\hline
		Data & \multicolumn{1}{c}{$K$} & \multicolumn{1}{c}{\gap GBT} & \multicolumn{1}{c}{\gap RID} \\
		\hline
		\parbox[t]{30mm}{\multirow{3}{*}{CCLE $EC50$ }} 
 & 5 & \textbf{ 0.35 } & 0.50 \\
& 10 & \textbf{ 0.22 } & 0.36 \\
& 15 & \textbf{ 0.13 } & 0.22 \\
& 20 & \textbf{ 0.07 } & 0.08 \\
		\hline
		\parbox[t]{30mm}{\multirow{3}{*}{CCLE $IC50$ }} 
 & 5 & \textbf{ 0.30 } & 0.45 \\
& 10 & \textbf{ 0.23 } & 0.30 \\
& 15 & \textbf{ 0.16 } & 0.17 \\
& 20 & { 0.13 } & \textbf{0.09} \\
		\hline
		\parbox[t]{30mm}{\multirow{3}{*}{CTRP $EC50$ }} 
 & 20 & \textbf{ 0.58 } & 1.11 \\
& 100 & \textbf{ 0.48 } & 1.91 \\
& 150 & \textbf{ 0.47 } & 1.79 \\
		\hline
		\parbox[t]{30mm}{\multirow{3}{*}{MovieLens 100K }} 
 & 20 & \textbf{ 0.06 } & 0.12 \\
& 100 & \textbf{ 0.05 } & 0.24 \\
& 150 & \textbf{ 0.05 } & 0.26 \\
		\hline
	\end{tabular}
	\caption{Mean squared error measure with various latent dimension $K$ parameters for CCLE $EC50$, CCLE $IC50$, CTRP $EC50$, and MovieLens 100K datasets.}
	\label{table:covnergence_mse_reporte}
\end{table}

\begin{table}[h]
	\begin{tabular}{llll}
		\hline
		Dataset        & $K=5$ & $K=10$ & $K=15$  \\ \hline
		CCLE $EC50$ & 0.0 & 4.44e-16  & 4.44e-16\\
		CCLE $IC50$ & \textbf{1.99e-2} & 4.44e-16  & 6.66e-16\\
		\hline \hline
		Dataset        & $K=20$ & $K=100$ & $K=150$  \\ \hline
		CTRP $EC50$& 4.44e-16 & \textbf{2.69e-2} & 1.99e-15\\
		MovieLens 100K & 2.22e-16 & 6.66e-16  & 8.88e-16\\
		\hline
	\end{tabular}
	\caption{Magnitude that exceeds 1 in randomized ID algorithm with various latent dimensions $K$ for different datasets.}
	\label{table:bid_magnitude_exceeds}
\end{table}

\subsection{Comparison with Randomized ID}

Comparative results for the proposed GBT model and the RID model on the four datasets are given in Figure~\ref{fig:error_bid_all} and Table~\ref{table:covnergence_mse_reporte}. 
We also observe that, in all experiments, the MSEs of the observed entries are worse than those of all the entries.
In the results of CCLE $EC50$ and CCLE $IC50$ datasets, the proposed GBT model recovers smaller MSE when the latent dimension is lower than 15 (for both the observed ones and all entries); However, when $K$ tends to be closer to the rank (24 in this case), the RID becomes better. 
Similar results can also be observed on the CTRP $EC50$ dataset whose rank is 545, and on the MovieLens 100K dataset whose rank is 943.
This demonstrates that our proposed models are more competitive when we need a low-rank ID approximation that can distill data to a great extent. 


For the proposed Bayesian GBT and GBTN models, in all circumstances, the maximal magnitude of the factored matrix $\bW$ is no greater than 1 because the prior densities in Bayesian ID models guarantee the magnitude constraint.
The randomized ID, on the other hand, finds a weaker interpolative decomposition with a maximal magnitude of no more than 2. Table~\ref{table:bid_magnitude_exceeds} shows the magnitudes of the randomized ID model for different datasets and different latent dimensions. We may notice that when $K=100$ on the CTRP $EC50$ dataset, the maximal magnitude is about $1.0269$. Though this is not a big issue in many applications, it has potential problems for other applications that have a high requirement for numerical stability. \citet{advani2021efficient} also reports that the randomized ID may have a magnitude larger than 167 making it less numerical stable.

\section{Conclusion}
The aim of this paper is to solve the numerical stability issue of the randomized algorithms in computing the low-rank ID approximation.
We propose a simple and computationally efficient algorithm that requires little extra computation and is easy to implement for interpolative decomposition. Overall, we show that the proposed GBT and GBTN models are versatile algorithms that have good convergence results and better reconstructive performances on both sparse and dense datasets, especially for low-rank approximation ID. 
GBT and GBTN are able to force the magnitude of the factored matrix to be no greater than 1 such that numerical stability is guaranteed.

\small

\bibliography{bib}
\bibliographystyle{sty}
\balance

\onecolumn
\appendix

\section{Derivation of Gibbs Sampler for Bayesian Interpolative Decomposition}\label{appendix:gbt_gbtn_derivation}
In this section, we give the derivation of the Gibbs sampler for the proposed Bayesian ID models. As shown in the main paper, the observed data $(m,n)$-th entry $a_{mn}$ of matrix $\bA$ is modeled using a Gaussian likelihood with variance $\sigma^2$ and mean given by the latent decomposition $\bx_m^\top\by_n$ (Eq.~\eqref{equation:idbid-per-example-loss}),
$$
\begin{aligned}
p(a_{mn} | \bx_m^\top\by_n, \sigma^2) &= \normal(a_{mn}|\bx_m^\top\by_n, \sigma^2);\\
p(\bA| \btheta) = \prod_{m,n=1}^{M,N}\normal \left(a_{mn}| (\bX\bY)_{mn}, \sigma^2 \right) &
= \prod_{m,n=1}^{M,N} \normal \left(a_{mn}| (\bX\bY)_{mn}, \tau^{-1} \right),
\end{aligned}
$$
where $\btheta=\{\bX,\bY,\sigma^2\}$ denotes all parameters in the model, $\sigma^2$ is the variance, $\tau^{-1}=\sigma^2$ is the precision,

\paragraph{Prior}
We choose a conjugate prior on the data variance, an inverse-Gamma distribution with shape $\alpha_\sigma$ and scale $\beta_\sigma$, 
$$
	p(\sigma^2 | \alpha_\sigma, \beta_\sigma) = \inversegammadist(\sigma^2 | \alpha_\sigma, \beta_\sigma).
$$
We assume further that the latent variable $y_{kl}$'s are drawn from a general-truncated-normal prior,
\begin{equation}\label{equation:rn_prior_bidd_append}
p(y_{kl} |\cdot) = \generaltruncatednormal(y_{kl} | \mu_{kl}, \tau_{kl}^{-1}, a=-1, b=1)=	 \frac{\sqrt{\frac{\tau_{kl}}{2\pi}} \exp \{-\frac{\tau_{kl}}{2}(y_{kl}-\mu_{kl})^2  \}  }{\Phi((b-\mu_{kl})\cdot \sqrt{\tau_{kl}})-\Phi((a-\mu_{kl})\cdot \sqrt{\tau_{kl}})} u(y_{kl} | a,b),
\end{equation}
where $u(x|a,b)$ is a step function with a value of 1 when $a\leq x\leq b$ and 0 otherwise.

Let $\br\in\{0,1\}^N$ be the state vector with each element indicating the type of the corresponding column. If $r_n=1$, then $\ba_n$ is a basis column; otherwise, $\ba_n$ is interpolated using the basis columns plus some error term. 

\paragraph{Hyperprior}
To further favor flexibility, we choose a convenient joint hyperprior density over the parameters $\{\mu_{kl}, \tau_{kl}\}$ of GTN prior in Eq.~\eqref{equation:rn_prior_bidd_append}, namely, the GTN-scaled-normal-Gamma (GTNSNG) prior,
\begin{equation}
	\begin{aligned}
		&\gap p(\mu_{kl}, \tau_{kl} |\cdot) = \gtnsng(\mu_{kl}, \tau_{kl}| \mu_\mu, (\tau_\mu)^{-1},\alpha_t, \beta_t)\\
		&=	\big\{\Phi((b-\mu_\mu)\cdot \sqrt{\tau_\mu})-\Phi((a-\mu_\mu)\cdot \sqrt{\tau_\mu})\big\}
		\cdot 
		\normal(\mu_{kl}| \mu_\mu, (\tau_\mu)^{-1}) \cdot \gammadist(\tau_{kl} | a_t, b_t).
	\end{aligned}
\end{equation}
This prior can decouple parameters $\mu_{kl}, \tau_{kl}$, and the posterior conditional densities of them are normal and Gamma respectively due to this convenient scale.

\paragraph{Posterior}
Following the graphical representation of the GBT (or the GBTN) model in Figure~\ref{fig:bmf_bids}, 
the conditional density of $y_{kl}^W$ can be derived,
$$
\begin{aligned}
	&\gap p(y_{kl} | \bA, \bX, \bY_{-kl}, \mu_{kl}, \tau_{kl}, \sigma^2) \propto  p(\bA|\bX,\bY, \sigma^2) \cdot p(y_{kl}|\mu_{kl}, \tau_{kl} )\\
	&=\prod_{i,j=1}^{M,N} \normal \left(a_{ij}| \bx_i^\top\by_j, \sigma^2 \right)\times
	\generaltruncatednormal(y_{kl} | \mu_{kl}, (\tau_{kl})^{-1},a=-1,b=1) \\
	&\propto
	\exp\left\{  
	-\frac{1}{2\sigma^2} \sum_{i,j=1}^{M,N} (a_{ij} - \bx_i^\top \by_j)^2
	\right\}
	\exp \{-\frac{\tau_{kl}}{2}(y_{kl}-\mu_{kl})^2  \}
	u(y_{kl} | a,b)\\
	&\propto
	\exp\left\{  
	-\frac{1}{2\sigma^2} \sum_{i}^{M} (a_{il} - \bx_i^\top \by_l)^2
	\right\}
	\exp \{-\frac{\tau}{2}(y_{kl}-\mu_{kl})^2  \}
	u(y_{kl} | a,b)\\
	&\propto
	\exp\left\{  
	-\frac{1}{2\sigma^2} \sum_{i}^{M} \bigg( x_{ik} ^2y_{kl }^2  + 2x_{ik} y_{kl } (\sum_{j\neq k}^{N}x_{ij} y_{jl}-a_{il})\bigg)
	\right\}
	\exp \{-\frac{\tau_{kl}}{2}(y_{kl}-\mu_{kl})^2  \}
	u(y_{kl} | a,b)\\
	&\propto
	\exp\left\{  
	-y_{kl }^2\big(\frac{\sum_{i}^{M}  x_{ik} ^2}{2\sigma^2}+\textcolor{black}{\frac{\tau_{kl}}{2}} \big)
	+y_{kl } 
	\underbrace{\bigg(\frac{1}{\sigma^2}  \sum_{i}^{M} x_{ik}  \big(a_{il}-\sum_{j\neq k}^{N}x_{ij}
		y_{jl}\big)
		+\textcolor{black}{\tau_{kl}\mu_{kl}}
		\bigg)}_{\textcolor{black}{\widetilde{\tau} \widetilde{\mu}}}
	\right\}
	u(y_{kl} | a,b)\\
	&\propto \normal(y_{kl}| \widetilde{\mu},( \widetilde{\tau})^{-1})u(y_{kl} | a,b) 
	\propto \generaltruncatednormal(y_{kl}| \widetilde{\mu},( \widetilde{\tau})^{-1}, a=-1,b=1),
\end{aligned}
$$
where again, for simplicity, we assume the rows of $\bX$ are denoted by $\bx_i$'s and columns of $\bY$ are denoted by $\by_j$'s, $\widetilde{\tau} =\frac{\sum_{i}^{M}  x_{ik} ^2}{\sigma^2} +\tau_{kl}$ is the posterior ``parent precision" of the general-truncated-normal distribution, and 
$$
\widetilde{\mu} = \bigg(\frac{1}{\sigma^2}  \sum_{i}^{M} x_{ik}  \big(a_{il}-\sum_{j\neq k}^{N}x_{ij}
y_{jl}\big)
+\textcolor{black}{\tau_{kl}\mu_{kl}}
\bigg) \big/ \widetilde{\tau}
$$
is the posterior ``parent mean" of the general-truncated-normal distribution.

Given the state vector $\br=[r_1,r_2, \ldots, r_N]^\top\in \real^N$, the relation between $\br$
and the index sets $J$ is simple; $J = J(\br) = \{n|r_n = 1\}_{n=1}^N$ and $I = I(\br) = \{n|r_n = 0\}_{n=1}^N$. A new value of state vector $\br$ is to select one index $j$ from index sets $J$ and another index $i$ from index sets $I$ (we note that $r_j=1$ and $r_i=0$ for the old state vector $\br$) such that 
$$
o_j = 
\frac{p(r_j=0, r_i=1|\bA,\sigma^2, \bY, \br_{-ji})}
{p(r_j=1, r_i=0|\bA,\sigma^2, \bY, \br_{-ji})}
=
\frac{p(r_j=0, r_i=1)}{p(r_j=1, r_i=0)}\times
\frac{p(\bA|\sigma^2, \bY, \br_{-ji}, r_j=0, r_i=1)}{p(\bA|\sigma^2, \bY, \br_{-ji}, r_j=1, r_i=0)},
$$
where $\br_{-ji}$ denotes all elements of $\br$ except $j$-th and $i$-th entries.
Trivially, we can set $p(r_j=0, r_i=1)=p(r_j=1, r_i=0)$. Then the full conditionally probability of $p(r_j=0, r_i=1|\bA,\sigma^2, \bY, \br_{-ji})$ can be calculated by 
$$
p(r_j=0, r_i=1|\bA,\sigma^2, \bY, \br_{-ji}) = \frac{o_j}{1+o_j}.
$$
Finally, the conditional density of $\sigma^2$ is an inverse-Gamma distribution by conjugacy,
$$
\begin{aligned}
	&\gap p(\sigma^2 | \bX, \bY, \bA)
	= \inversegammadist(\sigma^2 | \widetilde{\alpha_\sigma}, \widetilde{\beta_\sigma}),
\end{aligned}
$$
where $\widetilde{\alpha_\sigma} = \frac{MN}{2}+\alpha_\sigma$, 
$\widetilde{\beta_\sigma}=\frac{1}{2} \sum_{i,j=1}^{M,N}(a_{ij}-\bx_i^\top\by_j)^2+\beta_\sigma$.

\paragraph{Extra update for GBTN model}
Following the graphical representation of the GBTN model in Figure~\ref{fig:bmf_bids}, the conditional density of $\mu_{kl}$ can be derived,
$$
\begin{aligned}
&\gap p(\mu_{kl} | \tau_{kl}, \mu_\mu, \tau_\mu, a_t, b_t, y_{kl})
\propto \generaltruncatednormal(y_{kl} | \mu_{kl}, (\tau_{kl})^{-1}, a=-1, b=1)
\cdot \gtnsng(\mu_{kl}, \tau_{kl}| \mu_\mu, (\tau_\mu)^{-1},\alpha_t, \beta_t)\\
&\propto\generaltruncatednormal(y_{kl} | \mu_{kl}, (\tau_{kl})^{-1}, a=-1, b=1)
\cdot 
\big\{\Phi((b-\mu_\mu)\cdot \sqrt{\tau_\mu})-\Phi((a-\mu_\mu)\cdot \sqrt{\tau_\mu})\big\}
\cdot 
{\normal(\mu_{kl}| \mu_\mu, (\tau_\mu)^{-1})} \cdot 
\cancel{\gammadist(\tau_{kl} | a_t, b_t)}\\
&\propto 
\sqrt{\tau_{kl}}\cdot \exp\left\{ -\frac{\tau_{kl}}{2} (y_{kl}-\mu_{kl})^2\right\}
\cdot \exp\left\{ -\frac{\tau_\mu}{2}(\mu_\mu - \mu_{kl})^2  \right\}\\
&\propto \exp\left\{  -\frac{\tau_{kl}+\tau_\mu}{2} \mu_{kl}^2 + \mu_{kl}
\underbrace{(\tau_{kl}y_{kl}+\tau_\mu\mu_\mu)}_{\widetilde{m}\cdot \widetilde{t}}  \right\}\propto 
\normal(\mu_{kl}| \widetilde{m},(\,\widetilde{t}\,)^{-1}),
\end{aligned}
$$
where $\widetilde{t} = \tau_{kl}+\tau_\mu$, and $\widetilde{m} =(\tau_{kl}y_{kl}+\tau_\mu\mu_\mu)/\widetilde{t}$ are the posterior precision and mean of the normal density. Similarly, the conditional density of $\tau_{kl}$ is,
$$
\begin{aligned}
	&\gap p(\tau_{kl} | \mu_{kl}, \mu_\mu, \tau_\mu, a_t, b_t, y_{kl})
\propto \generaltruncatednormal(y_{kl} | \mu_{kl}, (\tau_{kl})^{-1}, a=-1, b=1)
\cdot \gtnsng(\mu_{kl}, \tau_{kl}| \mu_\mu, (\tau_\mu)^{-1},\alpha_t, \beta_t)\\
&\propto\generaltruncatednormal(y_{kl} | \mu_{kl}, (\tau_{kl})^{-1}, a=-1, b=1)
\cdot 
\big\{\Phi((b-\mu_\mu)\cdot \sqrt{\tau_\mu})-\Phi((a-\mu_\mu)\cdot \sqrt{\tau_\mu})\big\}
\cdot 
\cancel{\normal(\mu_{kl}| \mu_\mu, (\tau_\mu)^{-1})} \cdot 
{\gammadist(\tau_{kl} | a_t, b_t)}\\
&\propto \exp\left\{  -\tau_{kl}  \frac{(y_{kl}- \mu_{kl})^2}{2}  \right\}
\tau_{kl}^{1/2} \tau_{kl}^{a_t-1} \exp\left\{  -b_t \tau_{kl} \right\}
\propto \exp\left\{   -\tau_{kl}\left[ b_t +  \frac{(y_{kl}- \mu_{kl})^2}{2}  \right] \right\}
\cdot \tau_{kl}^{(a_t+1/2)-1}\\
&\propto \gammadist(\tau_{kl} | \widetilde{a}, \widetilde{b}),
\end{aligned}
$$
where $\widetilde{a} = a_t+1/2$ and $\widetilde{b}=b_t +  \frac{(y_{kl}- \mu_{kl})^2}{2}$ are the posterior parameters of the Gamma density.


\end{document}